# Training artificial neural networks to learn a nondeterministic game

Thomas E. Portegys

DigiPen Institute of Technology

9931 Willows Rd. NE, Redmond, WA, 98052 USA

portegys@gmail.com

**Abstract**.

It is well known that artificial neural networks (ANNs) can learn deterministic automata. Learning nondeterministic automata is another matter. This is important because much of the world is nondeterministic, taking the form of unpredictable or probabilistic events that must be acted upon. If ANNs are to engage such phenomena, then they must be able to learn how to deal with nondeterminism. In this project the game of Pong poses a nondeterministic environment. The learner is given an incomplete view of the game state and underlying deterministic physics, resulting in a nondeterministic game. Three models were trained and tested on the game: Mona, Elman, and Numenta's NuPIC.

**Keywords**: Mona artificial neural network, Elman artificial neural network, NuPIC hierarchical temporal memory, nondeterministic learning, game learning.

## 1 Introduction

Games, like many endeavors, are about reacting to and predicting events in the pursuit of goals. Games often also feature sequential actions as part of their play. Artificial neural networks (ANNs) have demonstrated considerable success in sequence prediction [1, 2]. For "conventional" multilayer perceptron (MLP) types of ANNs, sequences to be recognized are trained into the network beforehand as a set of static patterns. Because of this, reacting to untrained events is not a strength. This is an impediment to the use of MLPs in types of games that require this capability. Recent notable achievements playing arcade-style games [3] rely on the power of pattern classification rather than sequence recognition.

The aim of this project is to examine ANN architectures applied to learning a game that features both unpredictable events and sequential actions. These features are manifested in a nondeterministic finite automaton (NDA) [4]. Much of the world is nondeterministic, taking the form of unpredictable or probabilistic events that must be acted upon. If ANNs are to engage such phenomena, as biological networks do so readily, then they must be able to learn nondeterministic environments.

It is well known that recurrent MLPs, e.g. an Elman network, can learn deterministic finite automata [5, 6]. Learning the Reber Grammar is an example of this [7]. Learning nondeterministic finite automata is another matter. NDAs can produce event streams that are impossible to predict definitively, making comprehensive ANN training infeasible. This is anathema for MLPs that rely on such training to be effective.

In this project the game of Pong provides a nondeterministic environment. While a deterministic game of Pong can readily be learned by an ANN given the ball position and velocity [8, 9], in this project the learner is given an incomplete view of the game state and underlying physics, resulting in a nondeterministic game.

Three ANN models were trained and tested on the game:

- Mona, a goal-seeking network [10, 11].
- Elman, a popular MLP recurrent network [12].
- Numenta's NuPIC, a model of hierarchical temporal memory (HTM), which is closely based on neurological structure and function [13, 14].

## 2 Description

### 2.1 Pong game environment

The computer game of Pong [15] involves striking a moving ball with a movable paddle in a two-dimension playing area on a computer screen. Two paddles, controlled by opposing players, are positioned at the left and right ends of the playing area where they can be moved up and down to meet the ball. The ball can bounce off of the sides of the area as well as the paddles. A player loses when the ball gets by the player's paddle without being struck successfully.

In this project there is only one player, the ANN learner, controlling a paddle that is located on the right side of the playing area. A loss occurs when the ball passes the paddle, and a win is signified by a successful paddle hit. From the player's point of view, the playing area is overlaid by a 5x5 grid. The grid does not affect ball movements. The learner possesses sensors and response capabilities that are only effective in its currently located grid cell.

There are two sensors, one each for the ball and paddle states. Their values are supplied by the underlying game mechanics.

The ball sensor values:

BALL_ABSENT, BALL_PRESENT, BALL_MOVING_LEFT, BALL_MOVING_RIGHT, BALL_MOVING_UP, BALL_MOVING_DOWN

If the ball is moving up or down but also left or right, the ball sensor will report a vertical movement.

The paddle sensor values:

PADDLE_ABSENT, PADDLE_PRESENT

The learner can express these responses:

- CHECK_BALL: ask the physics for a ball sensor reading; this is only effective when the ball is in the current grid cell.
- TRACK_BALL_LEFT: move the learner's current grid cell left one cell, which is the correct response to the BALL_MOVING_LEFT sensor value.
- TRACK_BALL_RIGHT: move current grid one cell right.
- PAN_LEFT: move current grid cell left across the playing area until the ball or the left side is encountered.
- PAN_RIGHT: move current grid cell right until ball, paddle, or right side encountered.
- MOVE_PADDLE_UP: move paddle and current grid cell one cell up; this is only effective when the paddle is present.
- MOVE_PADDLE_DOWN: move paddle and current grid cell one cell down when the paddle is present.

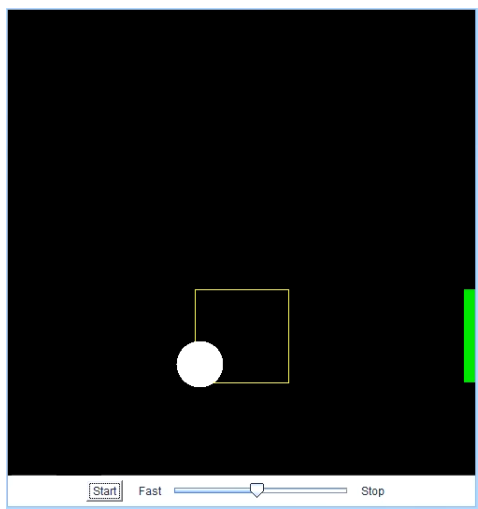

Figure 1 – Pong play. Box indicates sensory area.

Figure 1 is a snapshot of a game in progress. A video is also available on the web at http://youtu.be/Urdu9AJxoA0. Figure 2 shows the state space for winning games. The states are annotated with sensor states and the edges are annotated with responses. State transitions inputs are defined by implicit "step" signals which can have multiple target states, hence the state space embodies a nondeterministic automaton.

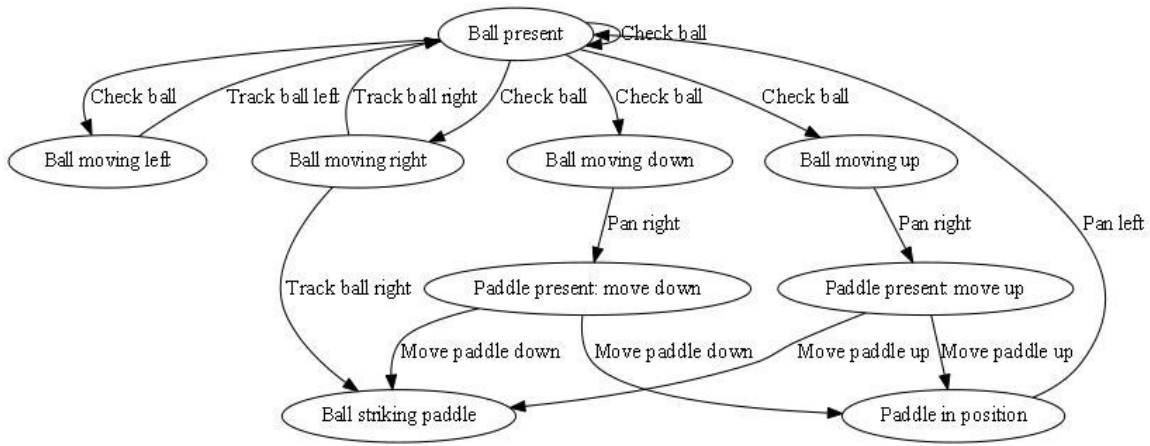

Figure 2 – Pong state space for winning games.

A game begins with the ball in the center of the playing area and the paddle in the center position. The ball is set to a random direction with a speed that is normalized so as not to outstrip the learner's ability to track it. Game play is turn-based, with the game mechanics changing the ball position and direction and the learner responding to sensor inputs.

**2.2 Training**

The learner is trained to track with the ball as it moves left and right. When sensors indicate that the ball is moving up or down, the learner is trained to (1) pan right to the paddle, (2) move the paddle up or down to remain aligned with the ball, and (3) pan left to locate the ball. The paddle-movement sequence is particularly challenging to train for two reasons: (1) the learner must remember which way to move the paddle without sensing the ball, and (2) the learner must remember that after it has moved the paddle and while continuing to sense the paddle, it must pan left to the ball.

**2.3 ANN models**

The following ANN models were trained and tested on the Pong game task. It should be mentioned that a fourth model, BECCA (Brain Emulating Cognitive Control Architecture) [16], was considered for comparison but was not included due to time constraints.

BECCA was exhibiting preliminary promising performance but was not optimally trained as of this writing.

### 2.3.1 Mona

Mona [10, 11] is a goal-seeking ANN that learns hierarchies of cause and effect contexts. These contexts allow Mona to predict and manipulate future events. The structure of the environment is modeled in long-term memory; the state of the environment is modeled in working memory. Mona uses environmental contexts to produce responses that navigate the environment toward goal events that satisfy internal needs. Because of its goal-seeking nature, Mona is also an example of reinforcement learning. Mona was selected for this task to illustrate the plasticity of a goal-seeking network in dealing with an NDA environment.

For the Pong task two sensors were configured, one for the ball and one for the paddle. A set of values between 0 and 1 were defined for the 6 ball sensor values and the 2 paddle sensor values. The response output ranged from 0 to 6 to encode the 7 possible response values.

### 2.3.2 Elman

An Elman network [12], also known as a Simple Recurrent Network, contains feedback units that allow it to retain temporal state information useful in classifying sequential input patterns. These feedback units reside in a context layer as shown in Figure 3. Each hidden layer unit has a connection to a corresponding context unit with a fixed weight of 1. An Elman network was selected for this task as a means of comparing non-MLP models with a popular MLP model.

For the Pong task the Elman network was created with Lens (Light efficient network simulator) [17]. The network was configured with 8 input units for the 6 ball sensor plus 2 paddle sensor values; 20 hidden and 20 context units; and 7 output units for the 7 possible response values. "Off"/"on" sensor values were 0/1. Outputs were similarly trained to values of 0 and 1. The learning rate was set to 0.2.

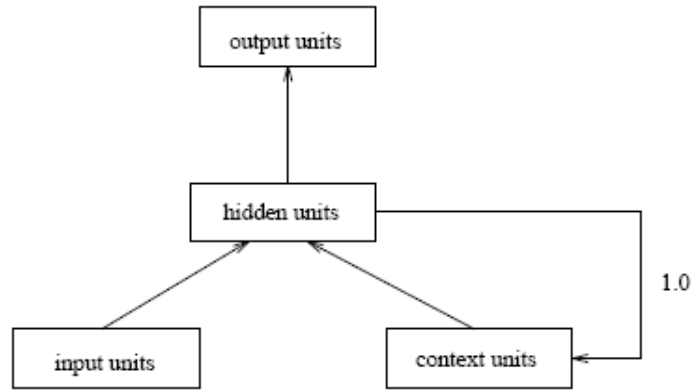

Figure 3 - Elman recurrent network.

### 2.3.3 NuPIC

NuPIC, the Numenta Platform for Intelligent Computing [13, 14], comprises a set of learning algorithms that attempts to faithfully capture how layers of neurons in the neocortex learn. NuPIC was selected for this task based on its successful performance in a number of sequential prediction tasks. At the heart of NuPIC is Hierarchal Temporal Memory, or HTM. From an algorithmic point of view there are three principle properties:

- Sparse Distributed Representations (SDRs): a sensor encoding technique that permits both noise tolerance and efficient pattern comparisons.
- Temporal inference: prediction of upcoming patterns in a stream.
- On-line learning: learning and prediction are concurrent.

For the Pong task the inputs and output were configured as they were for Mona.

## 3 Results

### 3.1 Training

For training, fifty random games of Pong were generated. For Mona, the BALL_PRESENT sensor state was defined as the only goal, which motivates the network to produce responses to navigate to the ball. As Figure 2 demonstrates, returning to the BALL_PRESENT state will generate winning Pong actions.

A single pass of the training set was given with the correct responses enforced on the network, and working memory cleared before each game. The learned Mona network is shown in Figures 4 and 5.

For the Elman network, 5000 training epochs were performed. For NuPIC, a swarm optimization using the training set was performed to select the optimal internal parameter values.

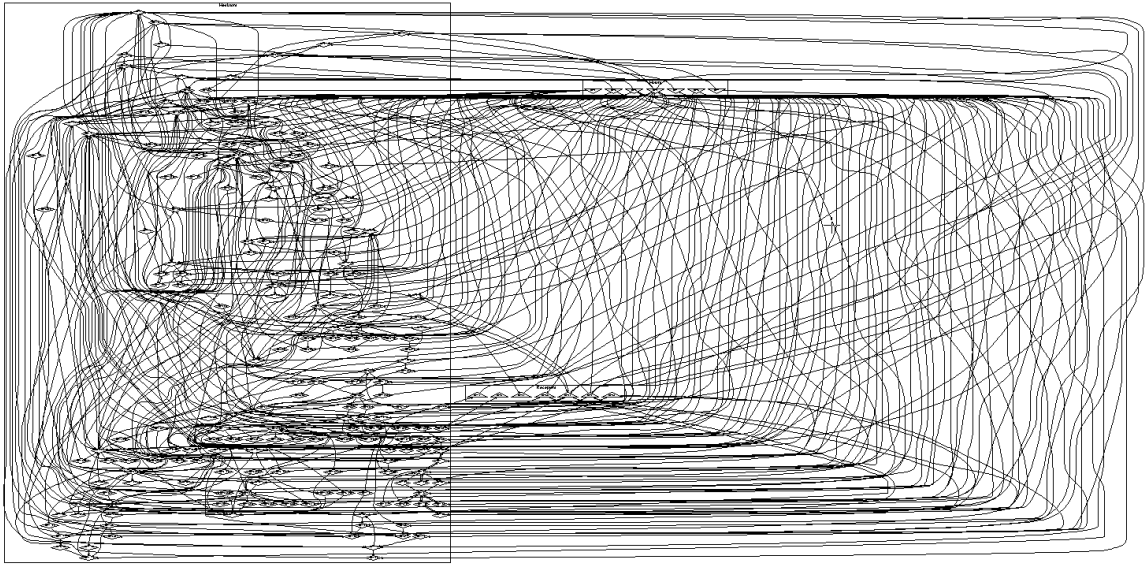

Figure 4 – Mona network after training.

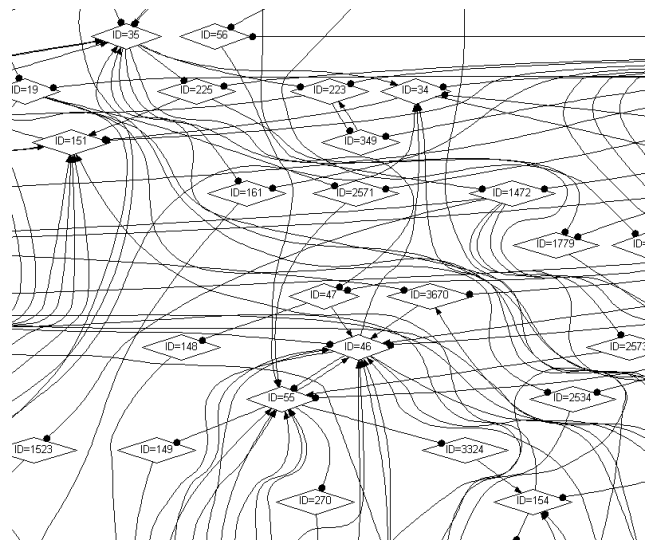

Figure 5 – Some mediator neurons in the trained Mona network.

## 3.2 Testing

For testing, a separate set of fifty random Pong games was generated. Each game was scored according to the percentage of initial consecutive correct responses toward winning the game. So if there were 10 responses to win a game and the learner output the first 8 correctly, the score for the game would be 80%. The rationale for this scoring scheme is that making any error will cast the learner off course from a winning response sequence.

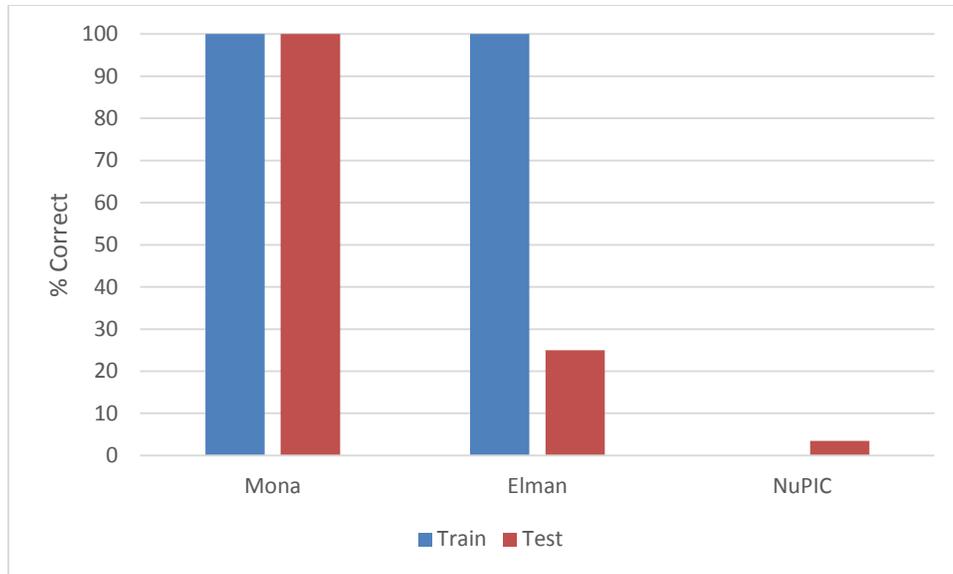

Figure 6 – Testing results.

Figure 6 shows results of testing with the test set as well as the training set for comparison (except for NuPIC). The Mona network performed perfectly for both sets. As might be expected, the Elman network performed perfectly for the set it was trained on, but much poorer for the test set, where it frequently encountered game play sequences that it was not trained to handle. This task obviously was not suitable for NuPIC, at least in its current form.

## 4 Conclusion

Even a simple nondeterministic game environment can pose significant problems for some ANN models, as the results show. For Mona, the goal-seeking component of its architecture is a major reason for its success on the task: it provides a mechanism for dynamically propagating motivation through a plethora of possible game sequences. For the Elman network, its success in predicting sequences that it has been trained on it notable. However, when sequences vary as they do in different untrained games, a marked decrease in performance ensues. For NuPIC, it seems clear that handling unpredictability is currently not a strong point. However, NuPIC remains under development as new neurological mechanisms are incorporated into it.

Modeling the brain has produced significant successes in the area of pattern classification for ANNs. However, the brain obviously has much more to teach in the domain of learning and executing behaviors that interact successfully with real-world environments. An aim of this project is to highlight the capabilities of models other than the prevalent multilayer perceptrons. These models can be complementary as well: for example, the pattern classification prowess of deep learning networks might be meshed with a behavioral oriented network such as Mona or a high fidelity neurological network model to form formidable hybrid architectures.

The open source code for Mona and the Pong project is available at http://mona.codeplex.com/ See the Readme in src/pong.